\documentclass{article}

\usepackage{PRIMEarxiv}

\usepackage[utf8]{inputenc} % allow utf-8 input
\usepackage[T1]{fontenc}    % use 8-bit T1 fonts
\usepackage{hyperref}       % hyperlinks
\usepackage{url}            % simple URL typesetting
\usepackage{booktabs}       % professional-quality tables
\usepackage{amsfonts}       % blackboard math symbols
\usepackage{nicefrac}       % compact symbols for 1/2, etc.
\usepackage{microtype}      % microtypography
\usepackage{makecell}
\usepackage{lipsum}
\usepackage{fancyhdr}       % header
\usepackage{graphicx}       % graphics
\graphicspath{{media/}}     % organize your images and other figures under media/ folder
\usepackage{tcolorbox} 
\usepackage{subfig}
\usepackage[frozencache=true,cachedir=minted-cache]{minted}
\title{Vision-Language Models and Automated Grading of Atopic Dermatitis
}

\author{
  Marc Lalonde, Hamed Ghodrati \\
  R\&D Dept. \\
  Computer Research Institute of Montreal (CRIM) \\
  405 Ogilvy Ave., \#101 \\
  Montreal, Qc, Canada\\
  \texttt{\{marc.lalonde, hamed.ghodrati\}@crim.ca}   
}

\begin{document}
\maketitle

\begin{abstract}
The task of grading atopic dermatitis (or AD, a form of eczema) from patient images is difficult even for trained dermatologists. Research on automating this task has progressed in recent years with the development of deep learning solutions; however, the rapid evolution of multimodal models and more specifically vision-language models (VLMs) opens the door to new possibilities in terms of explainable assessment of medical images, including dermatology. This report describes experiments carried out to evaluate the ability of seven VLMs to assess the severity of AD on a set of test images. 
\end{abstract}

% keywords can be removed

\section{Introduction}
\label{sec:intro}
Numerous efforts have been made to develop diagnostic support tools in medical imaging. The objective is to provide support to doctors (radiologists, cardiologists, ophthalmologists, etc.) to facilitate screening and enable them to focus on critical cases. This is also true in dermatology, particularly for the diagnosis of melanoma (skin cancer). Dermatology is a very broad field with more than 1,500 different types of diseases \cite{dermnet}. Since the advent of deep learning, many successful automated systems have been proposed, especially for the detection of melanoma \cite{BRINKER201911}. There is also a drive to develop generalist systems capable of detecting a wide range of diseases. A unique aspect of dermatology is that images are taken by handheld cameras operated by humans, unlike other subfields in medical imaging where sophisticated machines perform image acquisition.  These images can be of two types:

\begin{itemize}
\item Dermoscopic: Focused on the lesion, dermoscopic images are taken with a dermatoscope, which controls the illumination and the distance between the lens and the skin.
\item Clinical: Captured with a standard camera or a smartphone, these images include not only the lesions but also other body parts, clothing pieces, the background, etc.
\end{itemize}

The case of clinical images is challenging because automated diagnosis on such images requires that 1) the analysis algorithm searches for skin lesions within the image while ignoring distractors such as clothing and background, and 2) the algorithm deduces an overall diagnosis from localized diagnoses across the image (since the image may contain multiple lesion sites of varying severity). This scenario appears to be particularly suitable for new vision-language models (VLM) that can integrate textual information, such as the description of the injury, with visual information while considering spatial relationships of the data. These models are at the forefront of a wave of experiments in medical imaging \cite{saab2024capabilitiesgeminimodelsmedicine}, such as in radiology \cite{windsor2023visionlanguagemodellingradiologicalimaging}, cardiology \cite{echo}, ophthalmology \cite{holland2024specialistvisionlanguagemodelsclinical} and more.

In this report, we propose to study the performance of these VLM models in the specific task of grading a particular type of injury, atopic dermatitis (or AD, a form of eczema), in a clinical image. This task is difficult even for trained dermatologists \cite{Hurault2022-pv}. Dermatologists have developed and used several protocols called "grading tools", such as EASI and SCORAD, each with its own strengths and weaknesses \cite{RULLO2008205}. For example, the EASI grading consists of assigning a severity level between 0 and 3 to each of the following symptoms: 

 \begin{itemize}
\item erythema: skin redness;
\item induration or papulation: skin elevation (swelling);
\item excoriation: scratch mark;
\item lichenification: thickening of the skin with a leathery texture.
\end{itemize}

 The contributions of the report are as follows:
 \begin{itemize}
\item Investigate the ability of GPT-4o to perform an EASI grading on a set of test images with atopic dermatitis in zero-shot mode;
\item Compare its performance with some other commercial and open-source models;
\item Finetune two specific models (namely SkinGPT and PaliGemma) and assess their performance gain (if any). 
\end{itemize}

\section{Related Works}
\label{sec:works}

The early automated systems for diagnosing AD exploited deep learning, notably Convolutional Neural Networks (CNNs) to predict the severity scores. Bang et al. (2021) \cite{Bang2021} study the use of some CNN models for EASI grading on a large (20k images) dataset. The reported results are very high with an average accuracy of 95\%, possibly due to multiple factors such as the type of image used (close-up view of the lesions, similar to Fig. \ref{fig:easi-ug}), standardized camera conditions and uniqueness of the image source (Seoul St. Mary’s Hospital). In fact, the same authors report a \~10\% loss of accuracy when testing the models on data from another hospital. Similarly, Maulana et al. \cite{maulana2023evaluation} used five different CNN architectures including ResNet and EfficientNet \cite{tan2019efficientnet} to automatically score the severity of atopic dermatitis according to four qualitative levels: mild, moderate, severe, or none. They reported an accuracy of 89.8\% for ResNet-50 (the top performing architecture). Their dataset is, however, relatively small (\~3000 images) with low ethnic diversity (collected from 250 local patients); also, regions with skin lesions are manually cropped. The same team presented another pipeline in \cite{Suhendra_Suryadi_Husdayanti_Maulana_Noviandy_Sasmita_Subianto_Earlia_Niode_Idroes_2023} based on color feature extraction and Gradient Boosting. Their approach was tested on a portion of their dataset (500 images) for three factors, including erythema, lichenification, and pruritus (itching), achieving a total accuracy of 93.14\%. 

In a similar context, Pal et al. \cite{10.1007/978-3-030-01201-4_27} built three CNN architectures for the grading of psoriasis, one proposed by the authors, as well as ResNet \cite{He_2016_CVPR} and MobileNet \cite{howard2017mobilenets}. The grading is based on three indicators: erythema, induration, and scaling (silveryness), each classified into five ordinal severity levels: from 0 to 4. The models were trained and tested on a small dataset of 707 clinical images, manually cropped to minimize the presence of distractors such as hair, wrinkles, etc. PSENet \cite{Li_Wu_Zhao_Wu_Kuang_Yan_Ge_Wang_Fan_Chen_Wang_2020} incorporates CNNs into a Siamese architecture, analyzing a pair of images of skin lesions to estimate the severity of psoriasis based on erythema, induration, and scaling. Lesion attention modules that use the location of lesions determined by a pre-trained lesion detector are deployed into PSENet as well.

The advent of Large Language Models (LLMs) and, later on, Visual Language Models (VLMs) has revolutionized many fields including skin disease analysis, making it possible to have a conversation with an AI agent capable of analyzing a skin image. SkinGPT-4 \cite{zhou2024pre} is the fine-tuned version of MiniGPT-4 \cite{zhu2023minigpt}, trained on large amounts of skin disease images provided with doctors' notes. MiniGPT-4 is an open-source model that can be deployed locally, designed to integrate the capabilities of a large language model with visual information derived from a pre-trained vision encoder. The model leverages the open-source Llama as its language decoder and uses BLIP-2 as its visual encoder which consists of a vision transformer combined with a pre-trained Q-Former. SkinGPT-4 is trained in two steps, each with its own  dataset. The step 1 dataset is to familiarize the model with the medical features appearing in dermatological images without providing more details. The step 2 dataset is employed to make the model predict a more precise diagnosis using more detailed doctors' notes.     

\section{Description of the Evaluation Task}
\label{sec:task}

In this study, the VLM model is responsible for grading an image of a patient with atopic dermatitis. As the diagnosis is already known, the goal is to measure the ability of a VLM to assess the severity of AD according to the EASI scoring tool on a set of test images.
Each severity level is defined as follows for each of the four symptoms (erythema, induration/papulation, excoriation and lichenification):

\begin{itemize}
\item 0: None, absent
\item 1: Mild (just perceptible)
\item 2: Moderate (obvious)
\item 3: Severe
\end{itemize}

\begin{figure}
    \centering
    \includegraphics[width=0.65\linewidth]{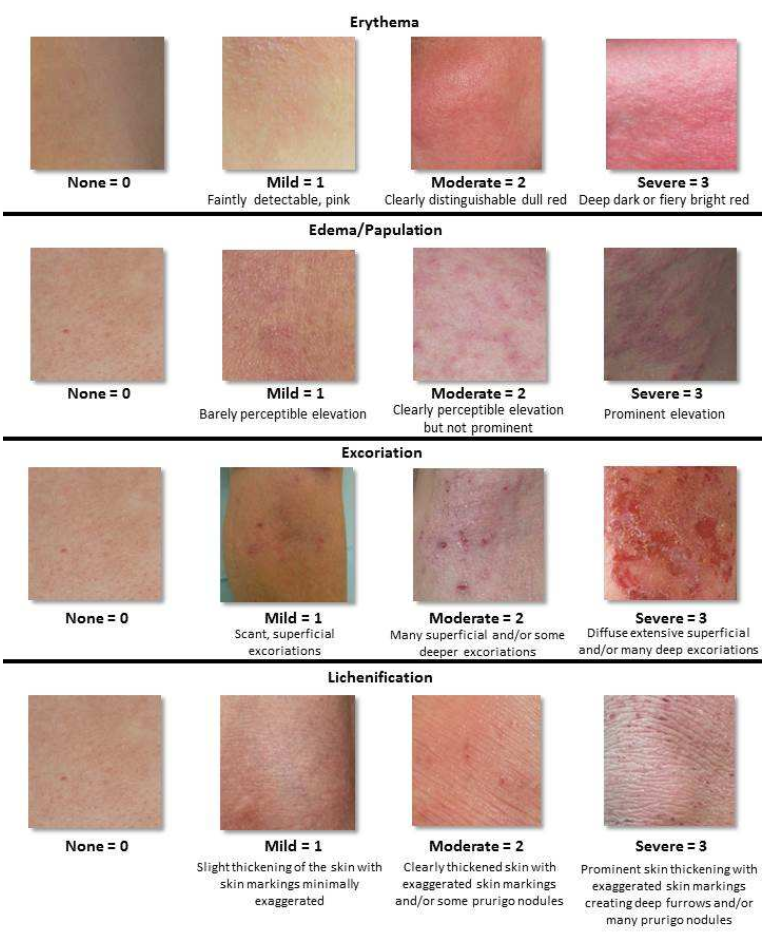}
    \caption{Visual examples of the EASI grading scale. From homeforeczema.org \cite{easi-ug}}
    \label{fig:easi-ug}
\end{figure}

Figure \ref{fig:easi-ug} visually illustrates the severity levels of each symptom.

\subsection{Performance Metrics}
The performance evaluation of a VLM varies according to each task. Although image captioning and some visual question-answering tasks require metrics based on the exact match between textual answers of the model and human-annotated answers (e.g. BLEU/ROUGE), the proposed performance evaluation strategy takes advantage of the numerical nature of the expected model output, which are severity levels on a [0-3] scale (EASI scoring). It then becomes easy to devise a metric that simply measures the difference between predicted and ground-truth levels for each symptom, provided that the model is prompted to provide the levels in a parsable form.

\subsection{Datasets}
\label{sec:datasets}
The availability of rich and unbiased datasets for the development of AI models applied to dermatology remains an issue even today. Huang et al.(2024) note in their survey on AI and eczema assessment \cite{HUANG2024102968} that the majority of the studies reported in the survey only use in-house images to train their respective models. If we exclude dermoscopic image datasets (since the objective is to address the more challenging application of clinical images), several datasets of interest can be considered (Table \ref{tab:dsets}). However, these datasets are not adequate for the proposed evaluation task: 1) the associated ground truth is often limited to presence/absence of a specific disease or, in the best cases, a very short textual description of the symptoms; 2) the number of images having DA as diagnosed pathology is small; and 3) no EASI-style annotation is provided, which is a hard limitation since the objective is to evaluate VLMs' ability to perform EASI gradings.

\begin{table}
    \centering
    \begin{tabular}{|l|c|l|}
    \hline
    Dataset & Statistics & Type of ground truth \\
    \hline\hline
        Fitzpatrick17k & 17 000 images & Many pathologies, \ few examples of eczema\\
         SKINCON & 3230 images from Fitzpatrick, all images from DDI & Presence/absence of 48 types of lesions \\
         DDI & 656 images & pathology + belign/malign label \\
         SCIN & +10 000 images & Many pathologies...\\
         SkinCAP & 4000 images (Fitzpatrick, DDI) & Diagnostic given in natural language\\
         \hline
    \end{tabular}
    \caption{Image datasets considered in this work}
    \label{tab:dsets}
\end{table}

The data preparation strategy followed these steps:
\begin{enumerate}
    \item Select SkinCAP \cite{zhou2024skincapmultimodaldermatologydataset} as the working dataset: the number of images is substantial, and, above all, a rich textual description is available for each image;
    \item Isolate images showing DA lesions only (many SkinCAP images are associated with two or more diagnoses, e.g. eczema combined with another disease; those were discarded);
    \item Ask a certified dermatologist to provide a EASI score for these DA images;
    \item Append to their corresponding in textual description a piece of text such as "In a four-level ordinal grading system, the overall severity assessment would be ? for erythema, ? for induration, ? for excoriation and ? for lichenification", with ? replaced with the appropriate EASI scores;
    \item From a teaching guide about DA and EASI grading \cite{easi-ug}, extract the sample images as well as their associated EASI scores and store them as a test dataset for model evaluation.
\end{enumerate}

There is an added benefit of constituting a test set based on sample images from this teaching guide compared to a subset of a public dermatology dataset: the probability of data leakage is minimized as these public datasets have probably been incorporated into the dataset used to train large models such as GPT-4o.  

\section{The Test Bench}

\subsection{Description}
The proposed test bench allows one to evaluate the AD grading capability of several VLM models using a simple command-line selection. It is also possible to provide the tool with a text file of results that will be evaluated with the same metrics as the methods implemented in the test bench (useful for evaluating methods that are difficult or impossible to integrate to the test bench, for instance closed-source models without public API).

\subsection{Included Models}
\label{inclmodels}

The following VLMs are accessible via the test bench: 
\begin{itemize}
    \item GPT-4o (via API): the de facto reference model due to its good performance in several areas.
    \item MiniCPM-o 2.6 (via Gradio API from a demo hosted on HuggingFace \cite{minicpm-hf}): open-source competitor to GPT-4o with good overall performance.
    \item PaliGemma2 (integrated): open-source model from Google, selected for its ease of fine-tuning
\end{itemize}
Integration efforts were also made for the InternVL2.6-o and SkinGPT models (Section \ref{sect:skingpt}), but the lack of documentation for their Gradio APIs \cite{gradio_api} prevented these efforts from being completed.  

The prompting strategy has been kept simple: a single prompt that requires the model to output its grading evaluation in the form of a Python list containing four digits between 0 and 3. Values correspond to erythema, induration, excoriation and lichenification respectively. The prompt, similar for all models, is as follows:
\begin{center}
\begin{tcolorbox}
[width=0.5\linewidth, sharp corners=all, colback=white!95!black]

Given that this is an image of a patient diagnosed with atopic dermatitis, estimate the severity score on a 0-3 scale, knowing that this score is a 4-digit python list corresponding to the degree of severity for erythema, papulation, excoriation 
and lichenification respectively.

\end{tcolorbox} 
\end{center}

\subsection{Usage}
The test bench is simple to use. It relies on the Hydra Python module \cite{hydra} for configuration management and is therefore easily configurable (via editing of the YAML parameter file or via the command line). For setup and preparation, a conda environment must be created and activated, in addition to updating the YAML file whose default values are adequate for typical use. The test bench configuration involves adding the API tokens required for access to the external services of OpenAI (GPT-4o), Kaggle (PaliGemma) and HuggingFace (MiniCPM-o 2.6). 

Start the testbed with the following command line:
%%\begin{tcolorbox}
%%[width=\linewidth, sharp corners=all, colback=white!95!black]

\begin{minted}
[
frame=lines,
framesep=2mm,
baselinestretch=1.2,
%%bgcolor=LightGray,
fontsize=\footnotesize,
]
{bash}
$ python main.py main.method=chatgpt
\end{minted}
%%\end{tcolorbox} 

At runtime, the specified method is evaluated on the test dataset described in Section \ref{sec:datasets} and the output produced includes two metrics: an MAE metric that reflects the average deviation, per symptom, between prediction and ground truth (viewed as a regression problem), and an accuracy score that shows how well a model can predict a symptom grading scale (viewed as a 4-class classification problem where each class corresponds to a scale level). 

 %%\begin{tcolorbox}
 %%[width=\linewidth, sharp corners=all, colback=white!95!black]
 %%\begin{verbatim}
\begin{minted}
[
frame=lines,
framesep=2mm,
baselinestretch=1.2,
%%bgcolor=LightGray,
fontsize=\footnotesize,
]
{bash} 
... 
=== gpt-4o === 
Averaging 5 trials 
5 images per trial 
Accuracies: 
  Erythema: 0.66 
  Induration: 0.52 
  Excoriation: 0.6 
  Lichenification: 0.62 
MAEs: 
  Erythema: 0.51 
  Induration: 0.55 
  Excoriation: 0.48 
  Lichenification: 0.44 
\end{minted}  
%%\end{verbatim}

%%\end{tcolorbox} 

More details are given in the README file accompanying the source code.

\section{Experiments}

\begin{figure}
  \centering
  \subfloat{\includegraphics[scale=0.35]{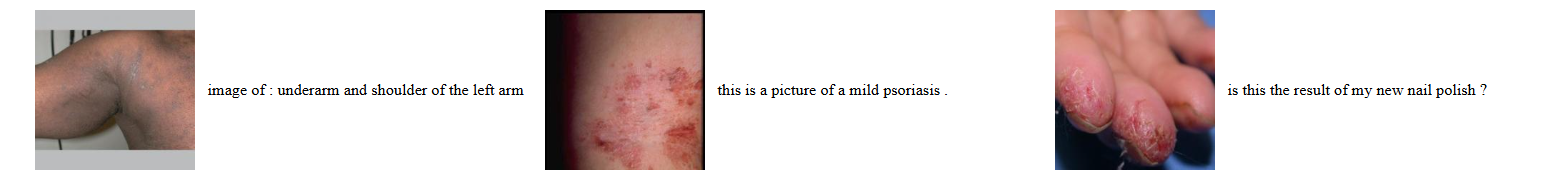} \label{fig:a}} \\
  \subfloat{\hspace*{0.15cm}\includegraphics[scale=0.365]{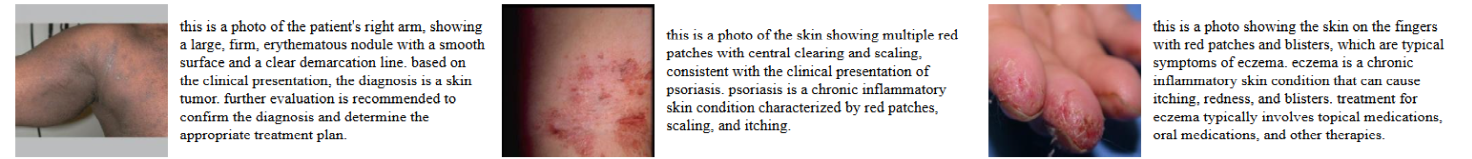} \label{fig:b}}
  \caption{Descriptions of some test images before and after fine-tuning PaliGemma~2} (drawn from the SkinCAP dataset \cite{zhou2024skincapmultimodaldermatologydataset}) \label{fig:befaft}
\end{figure}

\subsection{Experimenting with SkinGPT}
\label{sect:skingpt}
The SkinGPT demo code is implemented using Gradio, an interface that allows users to upload their images and chat with the model. As mentioned in Section \ref{inclmodels}, documentation issues with its Gradio API led us to modify the source code to allow batch processing of the test images. We first used the following two prompts:

\begin{tcolorbox}
 [width=\linewidth, sharp corners=all, colback=white!95!black]
 
\textbf{Prompt1:} 

\textit{Give me the severity score of the skin in this image for erythema, induration, excoriation, and lichenification. The scores must be 0, 1, 2, or 3 without any explanation.}\\
\end{tcolorbox}
\begin{tcolorbox}
 [width=\linewidth, sharp corners=all, colback=white!95!black]

\textbf{Prompt2:} 

\textit{Given that this is an image of a patient diagnosed with atopic dermatitis, estimate the severity score from 0, 1, 2, or 3 for the following symptoms: erythema, induration, excoriation, and lichenification. Only provide the score as a Python list.}
\end{tcolorbox}

However, we noticed that the model was unable to output any severity scores for either prompt. We asked ChatGPT-o1 to rewrite them:
\begin{tcolorbox}
 [width=\linewidth, sharp corners=all, colback=white!95!black]
\textbf{Prompt3:} 

\textit{You are a medical image analysis assistant. Please provide the severity scores (either 0, 1, 2, or 3) for each of the following factors:}
\begin{itemize}
    \item \textit{Erythema}
    \item \textit{Induration}
    \item \textit{Excoriation}
    \item \textit{Lichenification}
\end{itemize}
\textit{Give your response as four numbers only, in the order above, separated by commas. Do not provide any explanation.} \\
\end{tcolorbox}

\begin{tcolorbox}
 [width=\linewidth, sharp corners=all, colback=white!95!black]

\textbf{Prompt4:} 

\textit{You are a medical image analysis assistant. Assess the image for four factors: Erythema, Induration, Excoriation, and Lichenification. Provide an ordinal severity score between 0 and 3 for each. Return the scores in valid JSON with no additional explanation or text.}

\textit{Example format:}

\textit{\{}

\textit{"Erythema": 0}

\textit{"Induration": 0}

\textit{"Excoriation": 0}

\textit{"Lichenification": 0}

\textit{\}}
\end{tcolorbox}

Although these two prompts helped to reduce the number of non-compliant outputs, another approach was tested using a qualitative grading scale. The hypothesis behind this test was that the model would be more capable of handling textual keywords rather than numerical scales:
\begin{tcolorbox}
 [width=\linewidth, sharp corners=all, colback=white!95!black]
\textbf{Prompt5:} 

\textit{You are a medical image analysis assistant. Assess the image for four factors: Erythema, Induration, Excoriation, and Lichenification. Provide a qualitative severity score from normal, mild, moderate, or severe for each. Return the scores in valid JSON with no additional explanation or text.}
\end{tcolorbox}
 As the "non-compliance" issue still persisted, we modified the prompt to ask SkinGPT to only describe the issue with the skin, if any, with the output report being sent to chatGPT-o1 in order to predict the final ordinal severity scores. Here are the prompts to SkinGPT and ChatGPT, respectively:

\begin{tcolorbox}
 [width=\linewidth, sharp corners=all, colback=white!95!black]
\textbf{Prompt6:} 

\textit{You are an AI model specialized in describing skin images. For the attached images, please provide a structured, factual, non-numerical description of these four factors:} 

 \textit{Erythema: Color (e.g., faint pink, bright red), intensity (mild, moderate, intense), edges (well-defined vs. diffuse), and distribution.} 

 \textit{Induration: Presence or absence of thickening/firmness, any raised or swollen areas, and extent.}

 \textit{Excoriation: Presence or absence of scratch marks, erosions, or scabs, and overall distribution.}

 \textit{Lichenification: Degree of skin thickening or accentuated lines, extent, and any leathery appearance.}

 \textit{Do not offer diagnosis or treatment advice.} \\

 \textbf{ChatGPT prompt:} 
 
 \textit{You are an AI model specialized in grading skin images. Based on the following text, give me your estimation of the severity score from 0,1,2, or 3 in this order for these four factors: [erythema, induration, excoriation, lichenification]. No extra explanation! Only the scores}
\end{tcolorbox}

 Despite this prompt engineering effort, the fact that SkinGPT has not been trained to predict severity scores still causes non-compliant outputs to be rejected. We then fine-tuned the model using the enriched SkinCap dataset described in Section \ref{sec:datasets}. We tested the six prompts on the fine-tuned model and noticed an improvement on the rejection rate and also on the accuracy of the predicted scores. The best results obtained using prompt 6 are reported in Table \ref{tab:MAEs} and Table \ref{tab:acc}.  

\subsection{Test Bench Results}

This section reports the performance results collected with the test bench as well as some results obtained manually (via a web application). Globally, MAE errors (Tbl. \ref{tab:MAEs}) are somewhat high, especially for the smaller pretrained models, while  the larger models like GPT-4o,  Qwen2.5-VL-72b and InternVL2.5-78B seem to give better performance. Accuracy scores (Tbl. \ref{tab:acc}) follow the same path. Lower results for the grading of induration should not be a surprise since this symptom manifests itself as swelling, which is difficult to perceive in a 2D image. In general, performance is markedly lower compared to CNN methods found in the literature, which is expected from zero-shot methods. Fine-tuning of a powerful VLM model over a dataset with a decent size (thousands of AD images) would probably give much more interesting grading results with the valuable option of providing an explanation for the grading prediction. 

\subsection{Fine-Tuning PaliGemma}
An experiment has been carried out involving PaliGemma, a small open-source model proposed by Google \cite{paligemma}. Since, as their authors note, this family of models "require tuning in order to produce useful results", a fine-tuning step was performed on PaliGemma~2~3B using the enriched SkinCap data described in Section \ref{sec:datasets}. The expectation was that fine-tuning with these enriched descriptions would allow the model to develop a global understanding of skin disease terminology and, more specifically, the ability to grade AD images. After a 64-epoch training pass, textual outputs from some test set images clearly show that the model has gained some "dermatological knowledge" \cite{Lewandowski2024-ou} needed to describe a skin image (Fig. \ref{fig:befaft}). However, despite numerous prompt engineering attempts, it has not been possible to get the model to formulate an appropriate response: 1) guardrails regularly prevented the model from going further than a generic "patient should seek medical advice" message; 2) in situations where prompting would have induced a form of "grading-like" response, the model was incapable of generating a parsable output. In fact, a simple test with the prompt "Generate a list of four digits between 0 and 3" showed garbled output, which may indicate limitations in the expressive power of the LLM part (Gemma) of the model.

\begin{table}
    \centering
    \begin{tabular}{|c|c|c|c|c|c|}
    \hline
    VLM & \makecell{MAE\\(erythema)} & \makecell{MAE\\(induration)} & \makecell{MAE\\(excoriation)} & \makecell{MAE\\(lichenification)} \\
    \hline\hline
     GPT-4o & \textbf{0.43} & \textbf{0.56} & \textbf{0.46} & \textbf{0.4}  \\
     MiniCPM-2.6 8b & 1.08 & 1.08 & 1.125 & 0.58  \\
     InternVL2.5-78B & 0.5 & 0.625 & 0.58 & 0.625  \\
     Qwen2.5-VL 3b & 0.54 & 1.0 & 0.75 & 0.96  \\
     Qwen2.5-VL 72b & 0.5 & 0.75 & 0.96 & 0.58  \\
     Fine-tuned SkinGPT & 0.54 & 0.96 & 0.92 & 0.71  \\
%%     Fine-tuned PaliGemma2-3b & n/a & n/a & n/a & n/a  \\
     \hline    
    \end{tabular}
    \caption{MAE errors for all DA symptoms. Best score in bold. }
    \label{tab:MAEs}
\end{table}

\begin{table}
    \centering
    \begin{tabular}{|c|c|c|c|c|c|}
    \hline
    VLM & \makecell{Accuracy\\(erythema)} & \makecell{Accuracy\\(induration)} & \makecell{Accuracy\\(excoriation)} & \makecell{Accuracy\\(lichenification)} \\
    \hline\hline
     GPT-4o & \textbf{72\%} & \textbf{49\%} & \textbf{60\%} & \textbf{61\%}  \\
     MiniCPM-2.6 8b & 12\% & 29\% & 33\% & 54\%  \\
     InternVL2.5-78B & 71\% & 46\% & 46\% & 46\%  \\
     Qwen2.5-VL 3b & 62\% & 12\% & 33\% & 25\%  \\
     Qwen2.5-VL 72b & 71\% & 42\% & 29\% & 46\%  \\
     Fine-tuned SkinGPT & 67\% & 29\% & 38\% & 46\%  \\
     %%Fine-tuned PaliGemma2-3b & n/a & n/a & n/a & n/a  \\
     \hline    
    \end{tabular}
    \caption{Accuracy for all DA symptoms. Best score in bold.}
    \label{tab:acc}
\end{table}

\section{Discussion}

The reported evaluation results should be kept in perspective due to the test dataset used to measure performance. Not only does the limited number of images available for testing statistically restrict the interpretability of the results obtained, but since it did not allow for data curation, bias in the data is likely present (in terms of demographics, image quality, etc.). In addition, a more comprehensive test dataset should be annotated by more than one dermatologist. Yet this work should be seen as a first step toward understanding how VLMs can aid in automating the grading of atopic dermatitis. As for the next steps, quite a few directions are worth considering. Few-shot learning could be one, where examples are provided to the model in order to facilitate grading. Also, the experiments conducted during this work relied on the use of a single prompt submitted to each model; it would be interesting to explore whether chain-of-thought prompting can add value to the VLM-based grading process. Another direction of research is to investigate the potential benefit of hybrid architectures, i.e. combining VLM  models with e.g. Convolutional Neural Networks, which have some success in the automated grading of AD. This strategy might facilitate the use of smaller VLM models in a context where the use of (usually private) large models like GPT-4o is difficult to justify due to data privacy concerns.

\section*{Acknowledgments}
This was was supported in part by the Ministry of Economy, Innovation and Energy (MEIE) of the Government of Quebec. The authors wish to thank Zhuo Ran Cai, dermatologist at Centre Hospitalier de l'Université de Montréal (CHUM), for providing the EASI annotation for the test dataset.

%Bibliography
\bibliographystyle{unsrt}  
\bibliography{dermato}

\end{document}